# Evaluating Interactions between Automated Vehicles and Cyclists using a coupled In-the-Loop Test Environment

Michael Kaiser[1][0009-0005-7039-7239], Clemens Groß[1][0000-0002-1606-3845],
Lisa Marie Otto[1][0009-0002-5891-7316] and Steffen Müller[1][0000-0002-7831-7695]

[1] Department of Automotive Engineering, Technische Universität Berlin,
Gustav-Meyer-Allee 25, 13355 Berlin, Germany
michael.georg.kaiser@tu-berlin.de

**Abstract.** Testing and evaluating automated driving systems (ADS) in interactions with vulnerable road users (VRUs), such as cyclists, are essential for improving the safety of VRUs, but often lack realism. This paper presents and validates a coupled in-the-loop test environment that integrates a Cyclist-in-the-Loop test bench with a Vehicle-in-the-Loop test bench via a virtual environment (VE) developed in Unreal Engine 5. The setup enables closed-loop, bidirectional interaction between a real human cyclist and a real automated vehicle under safe and controllable conditions. The automated vehicle reacts to cyclist gestures via stimulated camera input, while the cyclist, riding a stationary bicycle, perceives and reacts to the vehicle in the VE in real time.

Validation experiments are conducted using a real automated shuttle bus with a track-and-follow function, performing three test maneuvers - straight-line driving with stop, circular track driving, and double lane change - on a proving ground and in the coupled in-the-loop test environment. The performance is evaluated by comparing the resulting vehicle trajectories in both environments. Additionally, the introduced latencies of individual components in the test setup are measured. The results demonstrate the feasibility of the approach and highlight its strengths and limitations for realistic ADS evaluation.

**Keywords:** Testing and Validation, Vehicle-in-the-Loop, Cyclist-in-the-Loop, Hardware-in-the-Loop, Automated Driving Systems, Mixed Reality, Motion Capture, Vehicle and Environment Interaction.

## 1 Introduction

Evaluating automated vehicles in interaction scenarios involving cyclists is particularly challenging due to safety concerns. Traditional testing methods, such as using moving dummy targets or relying solely on numerical simulations, fail to capture the authentic, dynamic behavior of real cyclists. Simulation-only environments are further constrained by model fidelity and the absence of real-time human behavioral feedback.

To study human cycling behavior, several cycling simulators have been developed in recent years (e.g., [1], [2]). However, these stand-alone simulators lack a feedback loop between the cyclist and an interacting vehicle, which is essential for investigating realistic interactions between traffic participants. Vehicle-in-the-Loop (ViL) test benches represent an advanced validation approach by integrating real vehicles into a closed-loop environment with virtual surroundings. For example, Dürr AG has developed a ViL test bench that combines a dynamic x-road roller system with over-the-air sensor stimulation, enabling real-time validation of camera- and radar-based ADAS functions [3]. In academic research, institutions such as KIT and TU Graz use ViL platforms to explore evaluation configurations for automated driving systems [4] and to conduct dynamic tests of highly automated functions in critical scenarios [5].

An alternative approach is being pursued at the MoSAIC VRU Laboratory of DLR [6], where multiple simulators (driver, pedestrian, and cyclist) are coupled to create a multi-user simulation environment. This setup enables investigations into cooperative driver assistance systems and the social dynamics of traffic interactions. However, it operates exclusively in the virtual domain; no real vehicles, physical sensors, control units, or actuators are integrated. As a result, hardware-level validation of automated vehicle behavior is not feasible in that framework.

To the best of the authors' knowledge, an approach that combines a Cyclist-in-the-Loop (CiL) and a ViL test bench has only been briefly proposed by the authors themselves [7], but has not yet been investigated in detail in the literature. This contribution addresses that gap by introducing such a test environment, evaluating how well the test bench replicates proving ground results, and assessing its potential for analyzing more complex interaction scenarios.

## 2    Coupled Test Benches System Architecture

The test setup presented in this work combines a ViL test bench with a CiL test bench via a shared virtual environment (VE), as illustrated in Fig. 1. The VE is implemented using Unreal Engine 5 in a local multiplayer configuration, where each test bench participates as a networked client. All clients operate at an update frequency of 90 Hz.

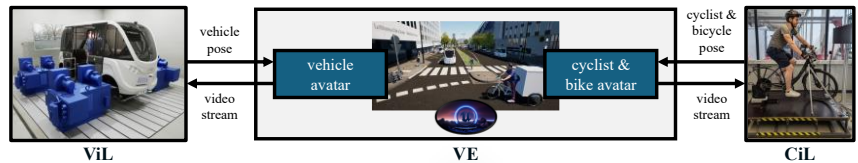

**Fig. 1.** Informational flow between ViL test bench, VE, and CiL test bench

This architecture enables bidirectional, closed-loop interaction: the automated vehicle can detect and respond to the cyclist's behavior via stimulated sensor input, while the cyclist perceives and reacts to the vehicle's motion in real time. The setup thus supports realistic, interactive testing scenarios between human cyclists and automated vehicles, without exposing the cyclist to any physical danger.

## 2.1  ViL Test Bench

The real automated vehicle on the ViL test bench perceives other traffic participants through camera stimulation. Visual content, rendered in Unreal Engine based on the vehicle's position and orientation, is projected at 60 Hz onto a screen in front of the vehicle's camera. The vehicle can respond by accelerating, braking, or steering against physically coupled actuators that simulate realistic force and torque feedback. Resistance torques are applied at each wheel by electrical motors to emulate longitudinal driving dynamics, while a steering actuator (Steering Force Emulator) provides dynamic counterforces at the steering rack. The vehicle's actions continuously update the state (e.g., position, velocity) of the vehicle's avatar in the VE. For more detailed information regarding the setup of the ViL test bench and the camera stimulation mechanism, see [8][9].

## 2.2  CiL Test Bench

The CiL test bench (see Fig. 2) enables a real human cyclist to interact with the VE while riding on a stationary bike trainer. The cyclist perceives the presence and behavior of other traffic participants on a screen, which displays visual content (field of view: horizontal 100°, vertical 37°) from the VE rendered in Unreal Engine based on the bicycle's position and orientation. The bike trainer is mounted on a mechanical platform that allows lateral tilting up to +/- 7° and applies a counter force via a mechanical spring mechanism, thereby simulating the balance dynamics of real-world cycling.

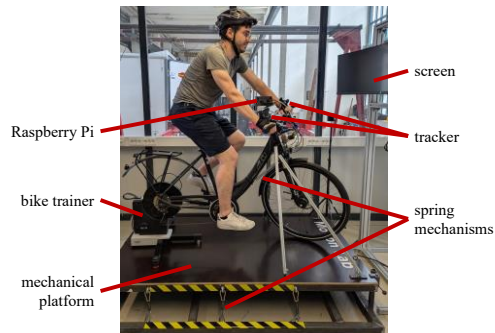

**Fig. 2.** Elements of the CiL test bench

The cyclist's behavior is captured through multiple sensors. Braking forces are measured using hydraulic pressure sensors embedded in the brake lines. The lean angle is recorded by an inertial measurement unit (IMU) fixed to the bicycle frame, and the steering angle is captured via a rotary potentiometer mounted on the headset star nut, aligned with the steering axis. Counter forces on the steering system are generated by a mechanical spring. Pedaling power is read from, and can be controlled via, the bike trainer over a Bluetooth connection. All sensor data are collected on a Raspberry Pi 3. There, they are aggregated and transmitted to the PC running the CiL's Unreal Engine instance via UDP over Ethernet. Additionally, hand positions are tracked using trackers attached to the cyclist's wrists and are directly sent to the same PC.

The behavior of the cyclist-and-bicycle avatar in the VE is continuously updated based on this input fed to a bicycle dynamics model based on [11]. A pedaling animation is shown based on the current bicycle velocity. Upper-body movement is derived from hand position via trackers while assuming a seated posture on the saddle; further joint angles are estimated using inverse kinematics to ensure biomechanical plausibility. The VE also computes resistance torque based on virtual driving resistances (e.g., gradient, aerodynamic drag, rolling resistance), which is then transmitted to the Raspberry Pi and used to configure the bike trainer.

## 3 Experimental Validation

To evaluate the real-world feasibility and performance of the proposed approach, we conduct a series of validation experiments.

### 3.1 Comparison between Proving Ground and Coupled Test Benches

To assess the performance of the coupled ViL-CiL test benches, we design a set of controlled experiments involving a real automated shuttle bus and a human cyclist. These are executed both on a proving ground (PG) and in the coupled test bench (TB) environment.

**Vehicle under Test and Track-and-Follow Function.** The vehicle under test (VUT), a real automated shuttle bus, is equipped with a simple track-and-follow function (TFF) designed to maintain a dynamic following behavior relative to a cyclist. This closed-loop setup allows the VUT to continuously adapt its trajectory in response to the cyclist's motion. The TFF serves as a functional demonstrator that enables comparison between the vehicle behavior in the physical environment and VE.

The TFF is implemented in ROS 2 [10] Humble, running on Ubuntu 22.04, and follows a modular sense-plan-act architecture. In the sense module, the VUT's front-facing, monocular First Sensor DC3K-1 camera (1280 x 960 px, 20 fps) provides image streams processed by a 3D human pose estimation algorithm to extract joint positions of detected individuals. The plan module identifies the nearest person as the tracking target. A raised hand above head level is interpreted as a start command; the same gesture is used to stop the function when repeated. To suppress oscillatory transitions between states, a hysteresis mechanism requires that the hand is detected below shoulder level before a subsequent command is accepted.

Once activated, a PID controller generates longitudinal velocity commands to maintain a desired following distance. Lateral control is handled via a steering controller that minimizes the yaw offset between the vehicle center axis and the current cyclist position. The act module transmits the resulting control commands to the vehicle's actuators via the CAN bus, using a drive-by-wire interface that incorporates conditional override logic to allow intervention capability for a safety operator.

**Test Scenarios.** To evaluate the coupled system, three maneuvers (see Fig. 3) are selected to cover a range of interaction types between the automated vehicle and the cyclist, including both longitudinal and lateral motion. Each maneuver begins with the cyclist in a predefined start position (B) and initiating the interaction by issuing a start gesture (hand raised above head). The test ends when the cyclist has reached a designated endpoint (E), the VUT has closed the gap, and a stop gesture is issued.

The test maneuvers are:
- Straight-line driving with intermediate stop: The cyclist rides along a straight path and stops at a predefined location (E'). Once the VUT catches up, the cyclist issues a stop command, moves forward a short distance (B'), issues a second start gesture, and completes the segment.
- Circular track driving: The cyclist follows a circular path with a radius of 16.5 m, completing 2¼ circuits.
- Double lane change: The cyclist performs a maneuver resembling a standard double lane change.

Each maneuver is executed at a target cyclist speed of approximately 4.5 km/h. The cyclist uses either a smartwatch (PG) or a velocity display on the CiL screen (TB) to adjust their speed.

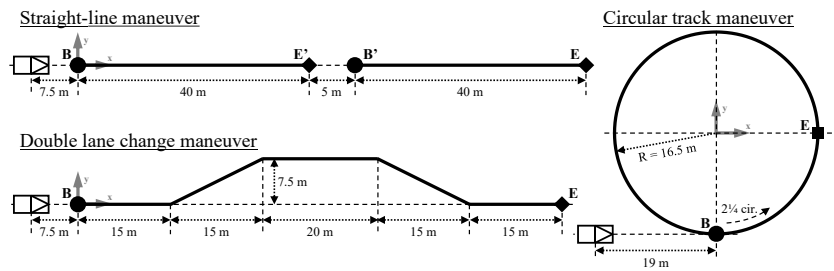

**Fig. 3.** Illustration of the conducted maneuvers

On the PG, cones are used to define the course geometry, ensuring consistent starting positions and heading angles across trials. The same course geometry is replicated in the VE, enabling the test person on the CiL test bench to follow the same path virtually. In both environments, the VUT operates the same real-time control system described above.

Each maneuver is repeated twice per environment, resulting in a total of 12 trials (3 scenarios × 2 environments × 2 repetitions). Data is recorded using ROS 2 in rosbag format. The VE provides a ROS 2 interface to deliver ground-truth data for all virtual participants.

In the PG scenarios, additional measurement data are recorded to support later evaluation, including signals from the TFF containing the images from the VUT's front camera as well as all available CAN bus messages and data from the VUT's additional onboard sensors: a Trimble BD892 GNSS receiver module, a vehicle-mounted VectorNav VN-100 IMU, and a front-facing Velodyne VLP-16 LiDAR.

**Test Evaluation and Results.** Data recorded in the TB environment contains full trajectory information in global coordinates, as this information is directly available from the VE during logging. In contrast, data collected in the PG scenarios require postprocessing to convert them into a comparable global format. To estimate the vehicle trajectory, we apply an extended Kalman filter that fuses CAN bus signals (wheel speeds and actual steering angle), GNSS data, and IMU measurements, employing a kinematic bicycle model to estimate the vehicle state - particularly its position and orientation - in global coordinates.

The cyclist's position is computed relative to the vehicle. For this purpose, we apply a YOLOv7 [12] object detector to images of the vehicle-mounted camera to identify the cyclist. LiDAR points are projected into the image plane, and a subset lying within the detector's 2D bounding box is selected to estimate the cyclist's relative position. By combining this estimate with the vehicle's global position and orientation, we derive the cyclist's absolute position in the PG scenario. We assume the cyclist is located at the maneuver starting point (B) at the moment the first start gesture is issued and transform both vehicle and cyclist positions accordingly, relative to the origin of the coordinate system.

Due to a misconfiguration of the scenario in the TB, the cyclist covers a shorter distance in both segments of the straight-line maneuver (see Fig. 4) compared to the PG scenario. However, the initial parts of each segment remain comparable. Taking into account the differences in cyclist speed, we observe very similar vehicle dynamics.

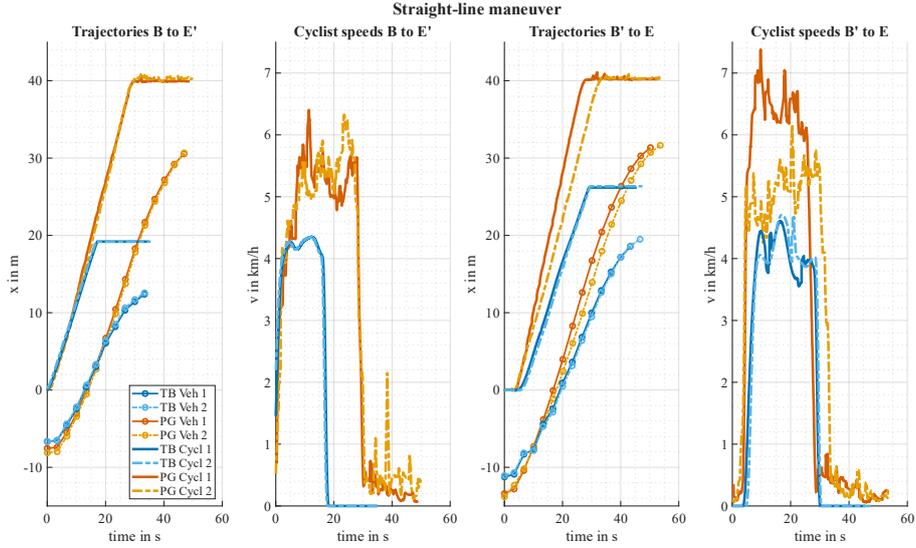

**Fig. 4.** Plots of trajectories and cyclist speeds for the straight-line maneuver

For the circular track maneuver (see Fig. 5), the vehicle's circular path is similarly replicated in both the PG and TB environments. Since the TFF generates control commands only based on the cyclist's current position (rather than, for example, the cyclist's traveled path), the cyclist speed strongly influences the vehicle's lateral control. At higher cyclist speeds, the vehicle's path radius becomes noticeably smaller. For similar cyclist speeds, we achieve comparable vehicle trajectories in both environments.

In the lane change maneuver (see Fig. 6), the cyclist successfully follows the target path in both longitudinal and lateral directions, indicating no significant limitations in steering or yaw control in either environment. Deviations in the vehicle's lateral position are likely caused by timing variations between issuing the start command and the cyclist's actual movement initiation. Again, we observe a strong influence of cyclist speed on the vehicle's lateral control. (Note that for readability, the y-axis scale in the trajectory plot for this maneuver is five times larger than the x-axis scale.) Taking these factors into account, the vehicle trajectories can be considered similar across both environments.

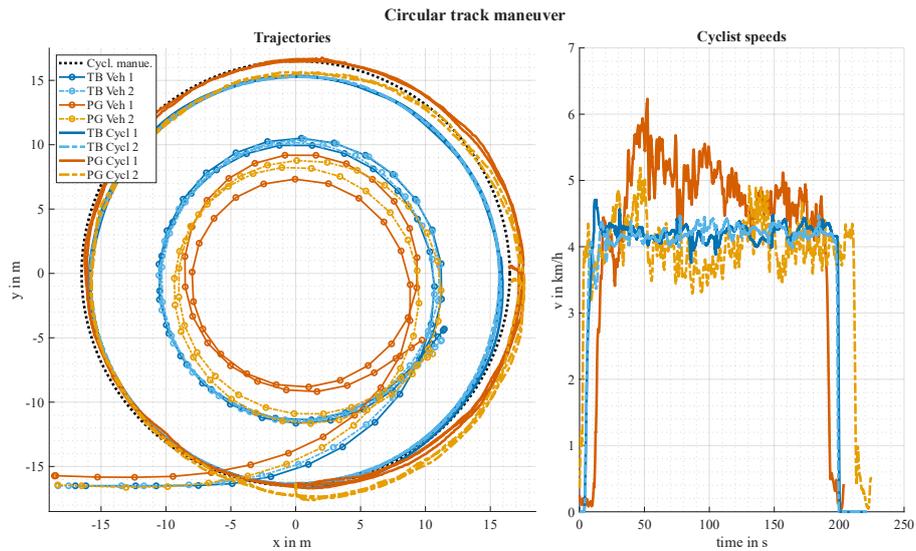

**Fig. 5.** Plots of trajectories and cyclist speeds for the circular track maneuver

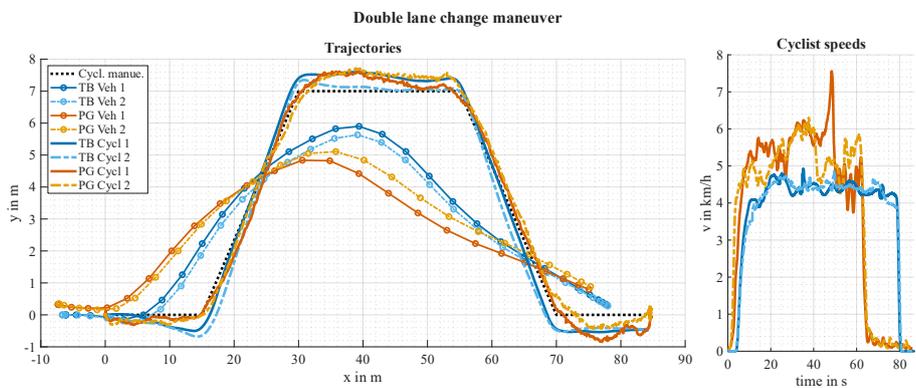

**Fig. 6.** Plots of trajectories and cyclist speeds for the double lane change maneuver

### 3.2  Latency Measurements

To estimate the latency introduced by the CiL test bench compared to PG tests, we record both the human subject in the CiL test bench and the camera stimulation image used by the vehicle in the ViL test bench. A high-speed camera operating at 240 frames per second is used, providing a temporal resolution of approximately 4.17 ms.

To assess the latency of human motion transmission through the CiL pipeline, the test subject repeatedly raises and lowers one arm in a semicircular upward or downward motion, which triggers corresponding movement of the cyclist avatar. Latency is measured as the number of video frames between the real arm reaching a 90° angle relative to the torso and the avatar's arm reaching the same position (see Fig. 7, left).

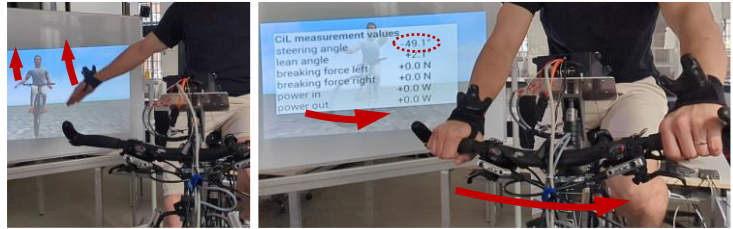

**Fig. 7.** Illustration of video frames during latency measurements - left: avatar only; right: including overlay with CiL sensor data (here: steering angle latency test)

Additional latency measurements are conducted for several cyclist-related variables: steering angle, lean angle, front and rear brake force, and pedaling power, under two conditions: (a) starting from standstill and (b) with a pre-spun flywheel on the bike trainer. For each modality, both the physical action and the corresponding camera stimulation image from the VE are recorded. Numeric values driving the avatar animation and internal state computations are overlaid on-screen to facilitate the analysis (see Fig. 7, right).

- Steering angle latency is measured by moving the handlebars in a sweeping motion and recording the time between the physical zero-crossing (centered position) and the avatar handlebars reaching the same orientation.
- Lean angle latency is evaluated by tilting the mechanical platform (and thus the bicycle) side to side, measuring the delay between the physical and the virtual zero-crossing positions.
- Brake latency is measured by actuating the respective brake lever and recording the time until a non-zero braking force is displayed.
- Pedaling power latency is measured by timing the interval between the onset of pedaling and the first non-zero power value displayed, with separate trials for stationary and pre-spun flywheel conditions.

For steering and lean angles - both directly observable in the avatar - we find no perceptible time discrepancy between the visual movement of the avatar and the displayed numerical values (e.g., steering angle vs. handlebar rotation).

Each measurement is repeated ten times. From these trials, we computed the mean latency, standard deviation, minimum, and maximum latency for each modality, as summarized in Table 1.

The avatar motion latency is lower than for the other measured modalities. We attribute this to the fact that the corresponding values are transmitted directly to the PC running the Unreal participant, bypassing the Raspberry Pi. Notably, we observe a substantial asymmetry in latency between upward and downward arm motions. This may result from the inverse kinematics solver used for avatar animation, which initializes from an A-pose and may inherently favor upward movements. However, further investigation is needed to confirm this and optimize the performance.

Most latency values fall within acceptable limits for real-time human-machine interaction. An exception is pedaling power, which exhibits delays exceeding one second. This delay appears to result from communication delays and internal processing within the commercial bike trainer hardware. Only minor differences are observed between the stationary and pre-spun flywheel conditions. Nonetheless, the impact on user immersion is expected to be limited: pedaling power primarily affects acceleration and only indirectly influences perceived velocity (via a first-order time integral) and position (via a second-order time integral), which attenuates the perceptual effect of latency.

Table 1. Latency measurements results. All values in milliseconds.

| modality | mean | std | min | max |
|---|---|---|---|---|
| avatar motion | 138 | 52.7 | 75 | 196 |
| (upward, downward) | (89, 188) | (11.3, 8.3) | (75, 175) | (100, 196) |
| steer | 225 | 23.2 | 188 | 263 |
| lean | 243 | 71.4 | 125 | 354 |
| brake force front | 198 | 18.3 | 158 | 217 |
| brake force rear | 210 | 24.4 | 167 | 250 |
| power (from standstill) | 1492 | 298.3 | 1163 | 2029 |
| power (from movement) | 1345 | 68.4 | 1204 | 1425 |

## 4 Summary and Outlook

This paper evaluated a test infrastructure that enables closed-loop, safe, and realistic interaction between a real automated vehicle and a real human cyclist. The system combines a CiL test bench and a ViL test bench via a shared VE, allowing both participants to interact in real time.

Experimental results from selected maneuvers demonstrated that the coupled test benches reproduce key interaction dynamics observed in proving ground scenarios. Vehicle trajectories remained consistent across both environments, proving the validity of the virtual setup for realistic interaction studies.

Latency measurements showed that most sensor and control signals are transmitted with sufficiently low delays to maintain real-time interactivity. An exception was observed for pedaling power, likely due to internal delays in the commercial bike trainer hardware. Additionally, asymmetric latency in avatar arm motion was noted between upward and downward gestures, potentially related to the inverse kinematics solver. These aspects have to be further analyzed.

Future work will focus on improving immersion and fidelity. Planned enhancements include the integration of spatial sound and the use of an additional tracker on the cyclist's helmet to capture head orientation and estimate viewing direction.

The proposed infrastructure provides a flexible, controllable, and safe platform for studying human-vehicle interaction in complex traffic scenarios, with the potential to improve both AV development and safety research.

## Acknowledgments

This work was partly funded by the German Federal Ministry for Economic Affairs and Climate Action (BMWK) and partly financed by the European Union in the frame of NextGenerationEU within the project "Solutions and Technologies for Automated Driving in Town" [13] (grant FKZ 19A22006P).

## References

1. Wintersberger, P., Matviienko, A., Schweidler, A., Michahelles, F.: Development and evaluation of a motion-based VR bicycle simulator. Proceedings of the ACM on Human-Computer Interaction 6(MHCI), pp. 1-19 (2022).
2. Bruzelius, F., Augusto, B.: CykelSim - Development and demonstration of an advanced bicycle simulator. Vinnova, Stockholm (2018).
3. Dürr AG: x-proof 360. Vehicle-in-the-Loop setup for ADAS or autonomous driving function tests using over-the-air stimulation, https://www.durr.com/en/products/final-assembly/testing-systems/autonomous-driving-passenger-cars/vehicle-in-the-loop-setup-for-adas, last accessed 2025/07/09
4. Fischer, A., Weber, Y., Freyer, J., Bause, K., Düser, T.: Evaluation of a Vehicle-in-the-Loop Test Bench Setup - Insights into a Systematic Validation Configuration Evaluation Approach. In: 2024 IEEE International Automated Vehicle Validation Conference, pp. 1–6. IEEE, Piscataway (2024).
5. Li, H., Makkapati, V.P., Wan, L., Tomasch, E., Hoschopf, H., Eichberger, A.: Validation of Automated Driving Function Based on the Apollo Platform: A Milestone for Simulation with Vehicle-in-the-Loop Testbed. Vehicles 5(2), 718–731 (2023).
6. German Aerospace Center (DLR), Institute of Transportation Systems: MoSAIC - Laboratory for multi-user simulation. https://www.dlr.de/en/ts/research-transfer/research-infrastructure/simulators/mosaic, last accessed 2025/07/09.
7. Kaiser, M., Otto, L.M., Müller, S., Hartwecker, A., Schyr, C.: Testing Urban Interaction Scenarios Between Automated Vehicles and Vulnerable Road Users Using a Vehicle-in-The-Loop Test Bench and a Motion Laboratory. In: Mastinu, G., Braghin, F., Cheli, F., Corno, M., Savaresi, S.M. (eds.) 16th International Symposium on Advanced Vehicle Control, pp. 791–797. Springer, Cham (2024).
8. Hartwecker, A., Müller, S., Schyr, C.: Safety of Use Analysis for Autonomous Driving Functions Under Laboratory Conditions. In: Orlova, A., Cole, D. (eds.) Advances in Dynamics of Vehicles on Roads and Tracks II, pp. 1183–1192. Springer, Cham (2022).
9. Otto, L.M., Kaiser, M., Seebacher, D., Müller, S.: Validation of AI-Based 3D Human Pose Estimation in a Cyber-Physical Environment. arXiv preprint arXiv:2506.23739 (2025).
10. Macenski, S., Foote, T., Gerkey, B., Lalancette, C., Woodall, W.: Robot Operating System 2: Design, architecture, and uses in the wild. Science Robotics 7(66), eabm6074 (2022).
11. Cossalter, V., Lot, R., Massaro, M.: Motorcycle Dynamics. In: Modelling, Simulation and Control of Two-Wheeled Vehicles, pp. 1–42. Wiley (2014).
12. Wang, C.-Y., Bochkovskiy, A., Liao, H.-Y.M.: YOLOv7: Trainable Bag-of-Freebies Sets New State-of-the-Art for Real-Time Object Detectors. In: 2023 IEEE/CVF Conference on Computer Vision and Pattern Recognition, pp. 7464–7475. IEEE, Vancouver (2023).
13. STADT:up Konsortium: Solutions and technologies for automated driving in town: an urban mobility project, https://www.stadtup-online.de/?language=en, last accessed 2025/07/09.